\documentclass[runningheads]{llncs}
\usepackage[T1]{fontenc}
\usepackage{graphicx}

\usepackage{multirow}
\usepackage{mathrsfs}
\usepackage[title]{appendix}
\usepackage{xcolor}
\usepackage{textcomp}
\usepackage{manyfoot}
\usepackage{booktabs}
\usepackage{algorithm}
\usepackage{algorithmicx}
\usepackage{algpseudocode}
\usepackage{listings}
\usepackage{tabulary}
\usepackage{siunitx}
\usepackage{multirow}
\usepackage{array}
\usepackage{subcaption}

\begin{document}
\title{Identification of Traditional Medicinal Plant Leaves Using an effective Deep Learning model and Self-Curated Dataset}
\titlerunning{Identification of Traditional Medicinal Plant Leaves}
%
\author{Deepjyoti Chetia\inst{1}\orcidID{0009-0004-7632-3700} \and
Dr. Sanjib Kr Kalita\inst{1}\orcidID{0000-0002-0576-3010} \and
Prof Partha Pratim Baruah\inst{1} \and
Debasish Dutta\inst{1}\orcidID{0000-0003-0315-3873} \and
Tanaz Akhter\inst{1}
}
\authorrunning{Deepjyoti et al.}
%
\institute{Gauhati University, Guwahati, Assam, India}
\maketitle              
\begin{abstract}
    Medicinal plants have been a key component in producing traditional and modern medicines, especially in the field of Ayurveda, an ancient Indian medical system.  Producing these medicines and collecting and extracting the right plant is a crucial step due to the visually similar nature of some plants. The extraction of these plants from non-medicinal plants requires human expert intervention. To solve the issue of accurate plant identification and reduce the need for a human expert in the collection process; employing computer vision methods will be efficient and beneficial. In this paper, we have proposed a model that solves such issues. The proposed model is a custom convolutional neural network (CNN) architecture with 6 convolution layers, max-pooling layers, and dense layers. The model was tested on three different datasets named Indian Medicinal Leaves Image Dataset,MED117 Medicinal Plant Leaf Dataset, and the self-curated dataset by the authors. The proposed model achieved respective accuracies of 99.5\%, 98.4\%, and 99.7\% using various optimizers including Adam, RMSprop, and SGD with momentum.


\end{abstract}


\section{Introduction}\label{sec1}

One of the world's recognized hotspots of biodiversity is North East India(NE), which houses a wide range of medicinal and aromatic plant species. This NE region comprises eight states (Assam, Arunachal Pradesh, Manipur, Meghalaya, Mizoram, Nagaland, Sikkim, and Tripura) with diverse ethnic communities having precious traditional knowledge and practices, passed through genesis. The pharmaceutical industry has made great use of the natural compounds and phytochemicals found in the medicinal plants of this region. People from the northeast, especially the tribal people, are ecosystem-friendly people who coexist peacefully with the natural world and uphold a strong bond between the environment and humans. The Indian subcontinent is home to more than 53.8 million tribal people who live in 5000 villages of tribal tribes that are dominated by forests. This area makes up 15\% of all Indian landmasses and is one of the greatest emporia of ethnobotanical wealth \cite{handique2009}. Out of India’s 427 tribal communities, the northeastern states—Arunachal Pradesh, Assam, Manipur, Meghalaya, Mizoram, Nagaland, Sikkim, and Tripura—are home to more than 130 important tribal communities (2001 census). Although there is currently no complete list of medicinal plants in North Eastern India, estimates suggest that the area is home to roughly 2000 different types of plants with medicinal uses \cite{handique2009}. On many accounts, traditional medical systems are still functioning. The use of medicinal plants as a cure for many illnesses has expanded as a result of the negative effects of several synthetic medications on human health. According to a recent survey conducted by the World Health Organization, around 80\% of the world's population uses herbal remedies. Additionally, the WHO asserts that a number of 21,000 distinct plant species have medicinal applications \cite{anand2019}. The present study aims to create a deep learning-based model that can be useful in distinguishing between the different classes of leaves of medicinal plants according to their leaf texture features. Common people can identify unknown species more quickly using these models. The diversity of both within and across species in plants makes it impossible to distinguish medicinal plant leaves from one another solely by subjective examination. As a result, there is a growing desire among researchers to create various computer-based models for the medicinal leaf/plant characterization process. In the proposed study, a deep learning-based model has been developed to classify medicinal leaves.

\subsection{Objective}\label{sec2}

In this study, a customized CNN model has been proposed for multiclass classification in three datasets, namely Indian Medicinal Leaves Image Datasets \cite{pushpagangadkar2023}, MED117 \cite{parismitasarma2023}, and a self-curated dataset. The images and the dataset have undergone preprocessing steps, mainly scaling, resizing, and data augmentation. Different optimizers such as SGD with momentum, RMSprop, and Adam at different epochs have been compared in the proposed model to gain optimum accuracy.

\subsection{Motivation}\label{sec4}

In the field of Ayurveda, an ancient Indian medical system, medicinal plants have played a significant role in the production of both traditional and contemporary medications. As several plants have similar visual characteristics, selecting and extracting the proper plant is an extremely important stage in the production of these medications. It takes human expertise to extract these plants from non-medical plants. By using computer vision techniques, the problem of precise plant identification will be resolved, and the requirement for a human expert in the collecting process will be minimized. Also, after the COVID-19 pandemic period, people are gradually inclined towards nature and natural products. Hence, they are now in search of the right way to identify medicinal plants before use and consumption. These problems can be solved by deep learning models and deep learning-based mobile applications. Such applications will help people identify a plant they want and provide a database to learn more about the plant and its uses. In this study, the authors have proposed a model that addresses these problems.

\subsection{Contribution}\label{sec3}
In this study, the authors have compared the custom CNN model using three datasets namely Indian Medicinal Leaves Image Datasets \cite{pushpagangadkar2023}, MED117 \cite{parismitasarma2023}, and a self-curated dataset.
The main contributions of the paper are listed below:
\begin{itemize}
    \item Curation of a medicinal plant dataset containing 42,250 images of 50 medicinal plants around the region of Assam, India.
    \item An improvised CNN architecture has been proposed.
    \item The proposed architecture achieved significantly high accuracy with a fixed and limited number of samples.
    \item Selecting state-of-the-art optimizers to reduce convergence time with high accuracy.

\end{itemize}

\subsection{Background}\label{sec5}
    \subsubsection{Convolution Neural Network(CNN)}\label{sub1sec5}
    CNN is a deep-learning algorithm that has significant dominance in the field of computer vision. A primary CNN model consists of a convolution layer for feature extraction, a pooling layer to downsample the feature map, An activation function to add non-linearity to the network, and a fully connected layer to take input from the previous layer and compute the final classification task. Along with these layers, the addition of features like stride, padding, and optimizers to minimize the loss function makes CNN a robust technique for computer vision tasks.

\section{Related work}\label{sub2sec5}
    Deep learning techniques have been utilized tremendously to solve problems in computer vision, natural language processing, speech processing, etc. Along with deep learning algorithms such as CNN and RNN, applying encoder-decoder and transfer learning methods has improved performance further.  The authors have reviewed several papers that employed deep learning techniques to address the challenge of medicinal plant identification.

    S. Roopashree et al. \cite{roopashree2021} curated a dataset named DeepHerb  that contains 2515 images of 40 plant types. The work concentrates on transfer learning using various pre-trained models and classification using ANN and SVM. The authors used VGG16, VGG19, InceptionV3, and Xception for feature extraction. The proposed model obtained 97.5\% accuracy. The authors also introduce a cross-platform mobile application for real-time identification of medicinal plants.

    K. Uma et al. \cite{uma2022} proposed a novel approach consisting of EfficientB4Net, a Convolutional Block Attention Module (CBAM), and a Residual Block Decoder. EfficientB4Net acts as an encoder on their architecture to encode input features and Residual Block Decoder reconstructs the encoded data to be as close to the original input as possible, by eliminating noise. Then encoded features from EfficientB4Net and global features from the CBAM are forwarded to a fully connected layer. The authors have used 20 medicinal plants that are found in southern India. Their proposed method has achieved 95\% accuracy.

    A. Uddin et al \cite{uddin2023} curated a dataset consisting of 5000 images of 10 medicinal plants found in Bangladesh . They applied pre-trained deep learning models VGG16, ResNet50, DenseNet201, InceptionV3, and Xception. Among the five models, DenseNet201 performed best with 85.28\% accuracy. Along with the pre-trained model they proposed and applied dense-residual–dense (DRD), dense-residual–ConvLSTM-dense (DRCD), and inception-residual–ConvLSTM-dense (IRCD) on their dataset. Among them, the DRCD model obtained an accuracy of 97\%. To improve classification accuracy further they adopted an ensemble approach with soft and hard ensembles. The hard ensemble obtained an accuracy of 98\%, while the soft ensemble achieved the highest accuracy of 99\%.

    S. Sachar et al \cite{sachar2023} took a transfer learning approach to identify medicinal plants . The authors have proposed a novel approach where they have extracted features using a neural network, pre-trained on ImageNet, and fed them into machine learning classifiers for prediction. The authors have prepared three different models and analyzed their performance. They have also proposed an ensemble approach based on stacking classifiers where a meta-classifier is trained using predictions of stacked classifiers. The authors have used benchmark datasets 'Swedish' and `Flavia'; their approach achieved an accuracy of 99.16\% and 98.13\% on both datasets.

    J. G. Thanikkal et al. \cite{thanikkal2023} proposed a shape descriptor algorithm for medicinal plant identification (SCAMPI) along with a mobile application to identify medicinal plants for normal people . The authors attempt to solve the problem of identifying immunity-boosting medicinal plants for regular people due to the similar nature of different plants. SDAMPI is a lossless algorithm that is based on freeman chain code and it accurately describes the leaf shape information. The authors of this paper have collected 1000 images of 10 ayurvedic medicinal plants from Kottiyoor village, Kannur District, Kerala State, India. Their model achieved 96\% and 87\% accuracy for images of dimensions 64x64 and 128x128 respectively.

    J. G. Thanikkal et al. \cite{thanikkal2023a} proposed a Deep - Morpho Algorithm that extracts and analyzes morphological features of medicinal plant leaves, venation, shapes, apices, and bases for the prediction of image class . They have collected more than 300 images from 18 different plant families and formed their dataset. Their model achieved 96\% accuracy on their deep learning model. The model was also tested on benchmark datasets 'Flavia', 'Swedish', and 'Leaf' and achieved 91\%, 87\%, and 91\% accuracy respectively.

    J. Miao et al \cite{miao2023} proposed a plant identification model for traditional Chinese herbs of 6 plant types. The dataset contains 7,853 images and image pre-processing such as normalization, grayscale image data, denoising, and enhancement were applied. Their proposed method is to use the ConvNext algorithm and add the ACMix algorithm to improve ConvNext. ACMix includes convolution operations and Multi-Head Self Attention operations. Their proposed model obtained 91.3\% accuracy.

    M. Sharma et al \cite{sharma2023} took two approaches for feature extraction . In the first approach, handcrafted feature descriptors characterized by edge histograms, oriented gradients, and binary patterns are proposed. In the second approach deep features are extracted by CNN using transfer learning is proposed. Their first approach attained accuracy 99\% accuracy. Their dataset consists of 1500 images of 40 medicinal plants.

    R. Azadnia et al \cite{azadnia2022} proposed an intelligent vision-based solution for medicinal plant identification. Their method is a custom CNN that includes a global average pooling(GAP) layer, a dense layer, a dropout layer, and a softmax layer. Instead of using a fully connected layer, they have opted for a GAP layer to prevent overfitting by reducing the number of parameters and complexity of the network.  Their dataset contains 750 images of 5 different plants. Their model achieved 99.3\% accuracy.

    M. A. Hajam et al \cite{hajam2023} Proposed an ensemble of pre-trained CNN hybrid models for the problem of medicinal plant identification . Their testing dataset is the Mendeley Medicinal Leaf Dataset which contains 1835 images of 30 plant classes. In this study, individual pre-trained models such as VGG16, VGG19, and DenseNet201 were trained and then the ensemble approach was used to combine these individual models. The Ensemble of VGG19+DensNet201 outperformed other models by achieving an accuracy of 99.12\%.

    Bhavya K.R et al. \cite{k.r2024a} present a novel approach to improving the accuracy and robustness of classifying medicinal plant leaf species. The study integrates ensemble learning techniques with a transfer learning approach. The authors have used pre-trained models VGG16 and  VGG19 to build the ensemble model. The dataset that is used by the author is Indian Medicinal Plant Leaves dataset [1]. The pre-trained models VGG16 and VGG19 were trained independently and obtained classification accuracy of 96.1\% and 97.6\% respectively. The ensemble model which consists of VGG16 and VGG19 models, is trained on the dataset and performs a classification accuracy of 99

    N. Rohith et al. \cite{rohith2023a} compare the performances of six different CNN models with attention mechanisms on the Indian Medicinal Plant Leaves dataset \cite{pushpagangadkar2023}. They have introduced the Efficient Channel Attention (ECA)  \cite{woo2018} module as a solution to the trade-off between performance enhancement and increased model complexity in deep convolutional neural networks and the Convolutional Block Attention Module (CBAM) \cite{woo2018} which is a lightweight and effective attention module for feed-forward convolutional neural networks. Among the six models, ECA-VGG19 achieved a top accuracy of 98.8\%.
    
    Sharma et al. \cite{SHARMA2024112147} introduces a novel multi-task learning network (MTJNet) that combines both local and global feature extraction to improve classification accuracy. MTJNet integrates the extracted features to enhance the contextual features. The authors has used indian Medicinal plant Leaves dataset \cite{pushpagangadkar2023} and to highlight significatnt features they applied sobel and vein morphometric analyses to images. Their network achieved accuracy of 99.71\%.

    A summary of related works is depicted in Table \ref{tab1}.

    \begin{table*}[h]
        \caption{Related works in medicinal plant identification}\label{tab1}
        
        \begin{tabular*}{\textwidth}{@{}p{2.5em} p{15em} p{17em} p{3em}@{}}
        \toprule
        Paper & Dataset & Techniques & Accuracy \\
        \midrule

        \cite{roopashree2021} & Own; DeepHerb(2515 images of 40 plant types) & Transfer learning with ANN, SVM classifiers & 97.5\% \\

        \cite{uma2022} & Own (20 plant types) & Encoder-Decoder network & 95\% \\

        \cite{uddin2023} & Own(5000 images of 10 plant types) & Transfer learning, Ensemble of the pre-trained network & 85.28\%, 99\% \\
 
        \cite{sachar2023} & Swedish, Flavia benchmark dataset & Ensemble stacked classifier & 99.16\%, 98.13\% \\
            
        \cite{thanikkal2023} & Own(1000 images of 10 plant types) & shape descriptor algorithm for medicinal plant identification (SDAMPI) along with a bigram-based DL algorithm & 96\% \\
            
        \cite{thanikkal2023a} & Own(300 images of 18 plant types), Flavia, Swedish, Leaf & DMA custom CNN & 96\%, 91\%, 87\%, 91\% \\
            
        \cite{miao2023} & Own(7,853 images of 6 categories) & ConvNext with ACMix algorithm & 91.3\% \\

        \cite{sharma2023} & Own(500 images of 40 plants) & Handcrafted feature descriptors and transfer learning on CNN pre-trained model & 99\% \\

        \cite{azadnia2022} & Own(750 images of 5 plants) & CNN with global average pooling layer & 99.3\% \\

        \cite{hajam2023} & Mendeley Medicinal Leaf Dataset (1835 images of 30 plants) & Ensamble of pre-trained model, VGG19+DensNet201 & 99.12\% \\
 
        \cite{k.r2024a} & Indian Medicinal Leaves Image Datasets(6900 images of 80 plants) \cite{pushpagangadkar2023} & VGG16, VGG19, Ensamble of VGG16 and VGG19 & 96.1\%, 97.6\%, 99\% \\
            
        \cite{rohith2023a} & Indian Medicinal Leaves Image Datasets \cite{pushpagangadkar2023} & ECA-VGG19 (Best model) & 98.8\% \\
        
        \cite{SHARMA2024112147} & Indian Medicinal Leaves Image Datasets \cite{pushpagangadkar2023} & MTJNet & 99.71\% \\
        
        \bottomrule

        \end{tabular*}
    \end{table*}

\section{Methodology}\label{sec6}
\begin{figure}[h!]
    \centering
    \includegraphics[width=0.6\textwidth]{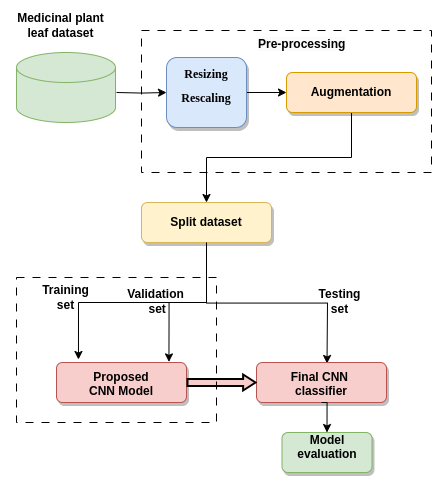}
    \caption{Proposed methodology}\label{fig1}
    \end{figure}

In this work, the identification of medicinal plant images and their multi-class classification through a convolutional neural network is proposed. Three datasets were used in this work. A customized CNN model was trained from scratch. Various optimizations and fine-tuning were done to improve the custom model's performance.  An overview of the proposed methodology is shown in Figure \ref{fig1} which shows the above-mentioned methodology.

\subsection{Dataset preparation}\label{sub1sec6}

\subsubsection{Data collection}
In this study, 3 datasets were used to evaluate the model's performance. Two of them were collected from publicly available domains and one was curated by the authors.

The first dataset that was selected for evaluation is curated by the authors. Images were collected from the Botanical Garden, Gauhati University, and Kaziranga Orchid Nursery in Bokakhat, Golaghat, Assam. All images in the data set were captured using an iPhone 14 Pro, a high-end smartphone known for its advanced camera system. The iPhone 14 Pro is equipped with several features that ensure high-quality image capture:

\begin{itemize}
	\item \textbf{Main Camera:} A 48 MP main sensor with advanced computational photography features.
	\item \textbf{Ultra-Wide Camera:} A 12 MP ultra-wide sensor for capturing more context.
	\item \textbf{Telephoto Camera:} A 12 MP telephoto lens with 3x optical zoom.
	\item \textbf{Li-DAR Scanner:} Enhances depth perception for better focus and AR applications.
\end{itemize}

The original resolution of images captured by the iPhone 14 Pro is 3024 x 4032 pixels. This high resolution allows for detailed capture of the leaf features, which is crucial for accurate plant identification and analysis. Images were captured by putting the leaves on white paper. Images are captured using different camera angles. Plant leaves were captured in daylight with illuminance range 32,000–100,000 lux. The dataset also has few blurred noisy images. These variations in image quality is common in real world scenarios. The dataset contains 50 medicinal plant types and no of images in each class ranges from 110 to 1800. During dataset preparation, the image dimension was reduced using a Python script. Hence leaf image dimension is set to 1024x720 and the image size is in the range of 20 KB to 110 KB. A subset of 10 plant classes is chosen for model evaluation. The total number of images in the subset is 11,289.

\begin{figure}[t!]
	\centering
	\includegraphics[width=0.7\textwidth]{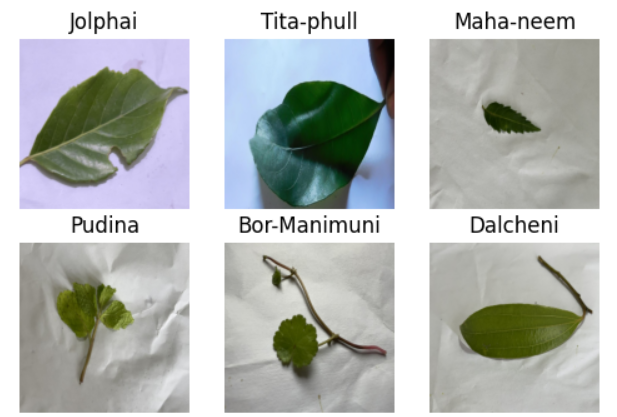}
	\caption{mages of the self-curated dataset}\label{fig2}
\end{figure}

The sample images of the classes of images present in the dataset are shown in figure \ref{fig2} .

The second dataset named Indian Medicinal Leaves image \cite{pushpagangadkar2023} was collected from the public domain Mendeley.com. The images are captured with varying backgrounds without any environmental constraints. The dataset consists of 80 types of plant leaf image classes and 40 types of plant image classes. Dimensions of images in this dataset are 450x650 pixels, image size varies between 50KB to 4.2 MB, and number of images in each class ranges between 8 to 177. Out of 80 plant types, a subset of 20 plant types were taken.  The total number of images in the subset is 1812. 

The third dataset named MED117 Medicinal Plant Leaf Dataset \cite{parismitasarma2023} was collected from the publicly available mendeley.com platform. This dataset consists of medicinal plant images that are collected from various parts of Assam, India. The dataset has two subfolders, a Raw leaf image set of medicinal plants and a Segmented leaf set. For this study, a raw leaf image set was used. The image dataset has 115 types of plant leaf classes with several images in each class ranging from 110 to 1800. Images are of dimension 1024x720 and image size ranges between 50KB and 500 KB. For evaluation, 10 classes were selected and the total no of images in the subset is 7341.

\subsubsection{Data augmentation}
\begin{figure}[t!]
	\centering
	\includegraphics[width=0.7\textwidth]{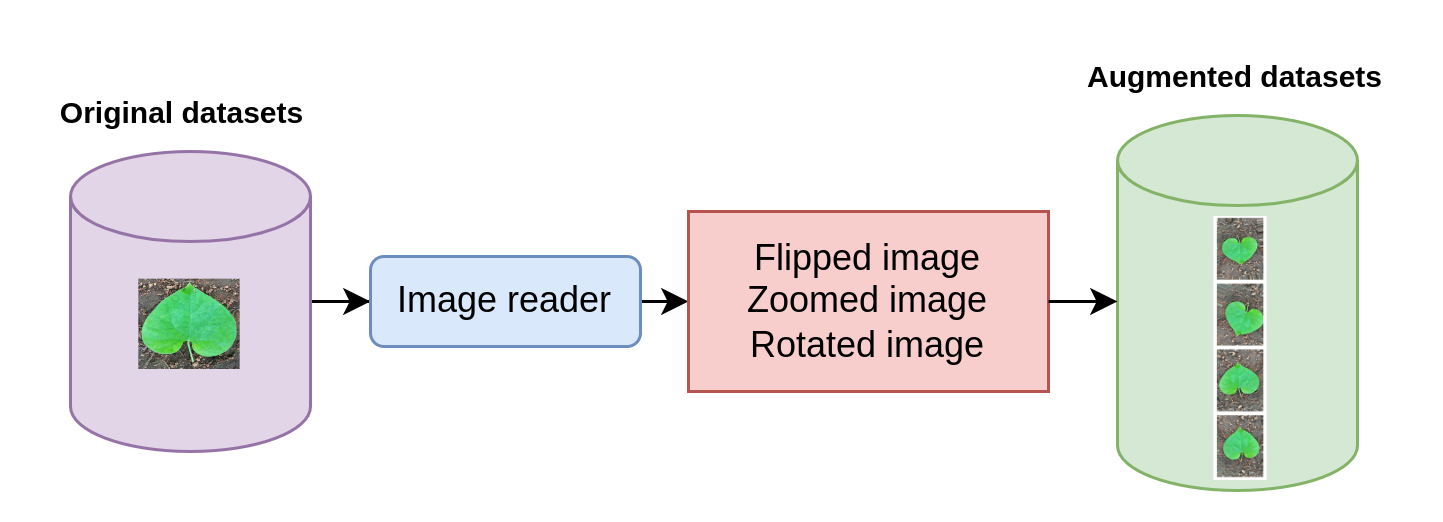}
	\caption{Image data augmentation process}\label{fig5}
\end{figure}

Data augmentation is applied to the dataset to avoid overfitting. For generating augmented images, augmentation techniques, i.e. random flip, random zoom, and random rotation, were applied. Images are flipped vertically and horizontally. All the images are zoomed in at 1.0. Medicinal plant leaf images are rotated at \ang{10} in an anti-clockwise manner. In this way, four images in total were generated from one pre-processed plant leaf image (figure \ref{fig5}).

\subsection{Data Preprocessing}\label{sub2sec6}
In preprocessing, a series of steps were taken including resizing, rescaling, and data augmentation.  Datasets are split into 80\% train, 10\% test, and 10\% validation set for our model training.

\subsubsection{Resizing}
In the collected datasets, medicinal plant leaf images are of different sizes. All the images are resized to 256x256 pixels, which are finally used for training, testing, and validation. Resizing images is a critical pre-processing step in computer vision. Principally, deep learning models train faster on small images. A larger input image requires the neural network to learn from four times as many pixels, and this increases the training time for the architecture.\cite{saponara2022}

\subsubsection{Rescaling}
RGB channel values of the input images are in the [0, 255] range. This is not ideal for a neural network. Input values were re-scaled to a range between 0 and 1. This is done by multiplying each cell by 1/255.	

\subsection{Proposed methodology}

\subsubsection{Model architecture}

\begin{figure}[h]
    \centering
    \includegraphics[width=13cm, height=3cm]{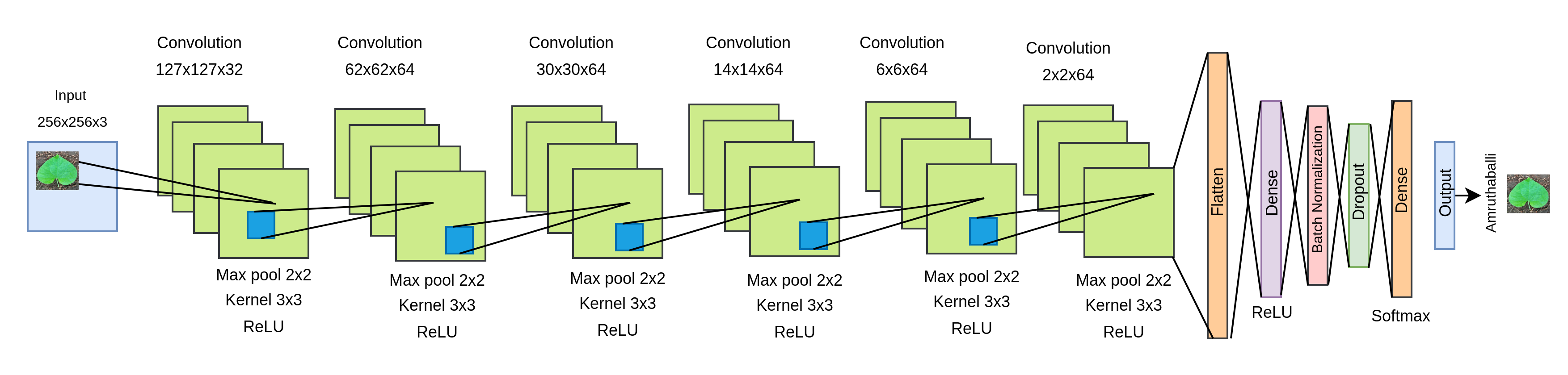}
    \caption{Proposed CNN architecture}\label{fig3}
    \end{figure}

After data pre-processing, images are fed into the proposed convolution neural network, shown in figure \ref{fig3}. The proposed model consisting of multiple convolutional and pooling layers to extract hierarchical features, followed by fully connected layer for classification. It contains 6 
convolution layers, 6 max-pooling layers, 2 dense layers. The proposed CNN model consists of a input layer that takes images of size 256x256 with 3 RGB channels and resizes and rescales the input images. It has six convolutional layers, each followed by a ReLU activation to capture progressively complex patterns, starting with 32 filters in the first layer and increasing to 64 filters in subsequent layers, all using 3x3 kernels. Each convolutional layer is followed by a max-pooling layer with a 2x2 pool size to reduce spatial dimensions and retain key features. After feature extraction, the output is flattened and fed into a fully connected layer with 64 units and ReLU activation, followed by batch normalization and a dropout layer (0.1) to improve generalization. The final layer is a softmax layer to output probabilities for each class. 

\subsubsection{Parameter Setting for the proposed model}

To update the different parameters, the training process of the proposed model has been evaluated iteratively. Some important parameters are batch size, regularization parameter, number of epochs, and learning rate, which determine the classification performance. One of the significant parameters influencing the model update phase is the batch size for deep learning. Several batch sizes were examined in the current study to train the model; ultimately, a value of 32 produced the best results. When there are more tunable parameters than samples in the training set, a deep neural network is more likely to overfit during training. Both BN and dropout have been employed in this scenario to establish regularization throughout training. In the present study, BN along with a 10\% dropout rate has been used to regularize the training process. Another hyperparameter epoch refers to the number of iterations the model passes through the training samples. In each iteration, every training sample in the training data set gets a chance to update the model parameter. It permits the learning algorithm to run until the optimal performance of the model is achieved. The number of epochs changes with different learning algorithms. The literature review reveals that there is no standard algorithm or mathematical model to set the value of the epoch. The best way to set the value of epochs is to gradually increase the value until the validation accuracy starts decreasing, even when the training accuracy increases. In the present study, we have tested the datasets using different values for epochs. Learning rate is another important hyperparameter that determines a value where the weights of the model have been adjusted with respect to the loss gradient. The smaller the learning rate, the slower the convergence; however higher learning rate may overshoot the solution region \cite{schmidt2021}. Therefore, the selection of the optimum learning rate plays an important role in achieving a good result. In our study, we have not set any learning rate, and the default value of 0.001 is considered for all the optimizers.

\subsubsection{Optimizer selection}

Due to the high processing cost associated with mathematical operations, optimizers are utilized during training to minimize the loss function and obtain the ideal network parameters in a reasonable amount of time \cite{li2022}. Optimization algorithms such as gradient descent try to minimize the error function by updating the weight vectors in a deep neural network. The gradient Descent algorithm faces some challenges like vanishing gradient problems, slow learning rate, etc. To solve these challenges, different optimization techniques have been proposed. These optimization algorithms aim to make gradient descent more efficient and faster. The proposed CNN model has been compared with different optimization algorithms to minimize the error function by updating the weight vectors. The proposed model has been computed with stochastic gradient descent(SGD) with momentum, root mean square propagation(RMSprop), and adaptive moment estimation(Adam) \cite{kingma2017}.

\section{Results and discussion}\label{sec7}

The proposed cnn model was tested on three different medicinal plant datasets, among which one was a self-curated dataset. The datasets are then given a 8:1:1 split for trainning, validation and testing. This model was trained using 3 different optimizers to check for and obtain a better accuracy. Also, hyperparameter batch size, image suze and epoch are varied along to test and obtain a high accuracy. Epoch values are gradually increased to attain the best possible results with minimal trainning.
To evaluate the performance accuracy, precision, recall, and f1 score metrics were used. Table \ref{tab2} depicts the results obtained after evaluating 3 datasets on the test split.

\begin{table}[h]
    \caption{Recorded evaluation metrics of 3 different datasets}\label{tab2}
    \begin{center}
    	\begin{tabulary}{\textwidth}{L L L L L}
    		\toprule
    		Dataset & Accuracy & Precision & Recall & F1 score \\
    		\midrule
    		
    		Indian medicinal plant \cite{pushpagangadkar2023} & 0.9955 & 0.9958 & 0.9955 & 0.9955 \\
    		
    		MED 117\cite{parismitasarma2023} & 0.9849 & 0.9871 & 0.9844 & 0.9848 \\
    		
    		self curated & 0.9974 & 0.9952 & 0.9951 & 0.9951 \\
    		
    		\bottomrule
    	\end{tabulary}
    \end{center}
\end{table}

\begin{figure*}[tp]
	\centering
	\begin{subfigure}{0.45\textwidth}
		\centering
		\includegraphics[width=\textwidth]{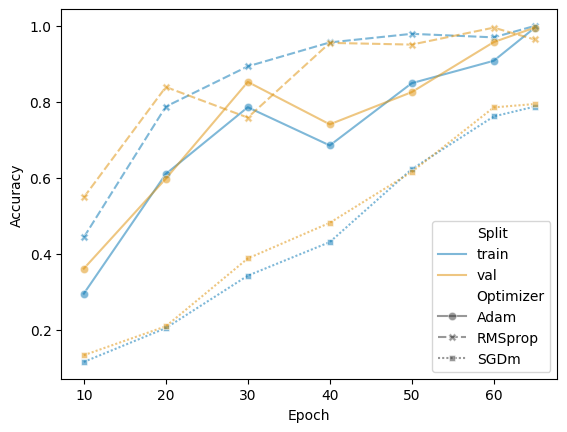}
		\caption{Indian medicinal plant dataset}
		\label{p1}
	\end{subfigure}
	\hfill
	\begin{subfigure}{0.45\textwidth}
		\centering
		\includegraphics[width=\textwidth]{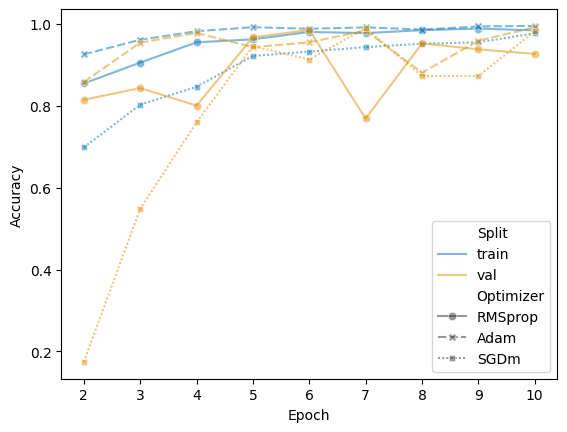}
		\caption{MED117 dataset}
		\label{p2}
	\end{subfigure}
	\hfill
	\begin{subfigure}[b]{0.5\textwidth}
		\centering
		\includegraphics[width=\textwidth]{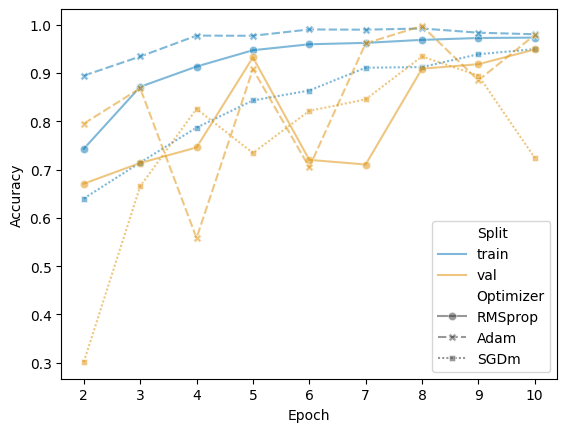}
		\caption{Curated dataset}
		\label{p3}
	\end{subfigure}
	\caption{Train and Validation accuracy vs. epoch of the model on three datasets}
	\label{res}
\end{figure*}

Figure \ref{res} illustrates the accuracy of the model while training and validation along each eopch on the 3 datasets. From the results, it is observed that the model achived highest accuracy 99.74\%, is on self-curated dataset with the Adam optimizer on the $8^{th}$ epoch. On the MED 117 dataset, highest accuracy was obtained simultaneously in $65^{th}$ epoch by Adam and $60^{th}$ by RMSprop. And the self curated dataset outperformed the Indian medicinal plant dataset on all optimizers while training with similar number of epochs.  

Also, amongst the models trained on the Indian medicinal plant\cite{pushpagangadkar2023}, the proposed model achived accuracy comparable to the best one. A table depicting the accuracies on different models is shown in Table \ref{tab3}

		\begin{table}[]
			\caption{Comparison with different models}\label{tab3}
			\begin{center}
				\begin{tabular}{ll}
					\toprule
					Model & Accuracy \\
					\midrule
					VGG16 \cite{k.r2024a}                            & 96.10\%  \\
					VGG19 \cite{k.r2024a}                            & 97.60\%  \\
					Ensamble VGG16 and VGG19 \cite{k.r2024a}         & 99\%     \\
					ECA-VGG19 \cite{rohith2023a}                        & 98.80\%  \\
					
					MTJNet \cite{SHARMA2024112147} &  99.71\% \\
					Proposed                            & 99\%   \\
					\bottomrule
				\end{tabular}
			\end{center}
			
		\end{table}

From the experiments, it can be observed that the performance of the proposed model increases with the size of the dataset. Datasets such as MED117 \cite{parismitasarma2023} and self-curated dataset with larger no. of image samples have shown good accuracy results in less no. of iteration. The three datasets that were used contain diverse medicinal plant leaf images.  Testing results of the model shown in Table \ref{tab2}, evaluated on the three different datasets indicate the scale of generalizability of the proposed model. Three optimizer algorithms were used to reduce the error while training the model, among them RMSprop showed the best results. Right after Adam, RMSprop and SGD (Momentum) performed best.
\section{Future work}\label{sec8}
\begin{itemize}
	\item To reduce convergence time and improve accuracy, the present work may be extended by utilizing advanced deep learning techniques such as encoder-decoder networks, and vision transformers. Expanding the dataset with more medicinal plant leaf images and increasing the no. of images per class could also improve accuracy and convergence time. 
	
	\item To reduce training time, a transfer learning approach can be taken. Transfer learning also solves the issue of a lack of high-end computing resources by using pre-trained models.
	
	\item To evaluate the model in real world scenarios, mobile application for medicinal plant leaf identification can be developed.  
\end{itemize}

\section{Conclusion}\label{sec9}

In this article, a custom deep-learning CNN model created from scratch is used for the identification of medicinal plants. Three datasets were used for the evaluation of the performance of our model. After applying image pre-processing steps, the model was trained with different optimizers, such as SGD with momentum, Adam, and RMSprop on the datasets. It is observed that Adam and RMSprop have performed best in all three datasets. The proposed CNN model along with RMSprop or Adam as optimizers could be an effective solution in the field of medicinal plant identification. 

\section{Declarations}\label{sec10}

\subsection{Data Availability Statement}

\begin{itemize}
	\item Indian medicinal leaves dataset\cite{pushpagangadkar2023} can be accessed through hyperlink \\ 
	(\url{https://data.mendeley.com/datasets/748f8jkphb/3}).
	
	\item MED117 dataset \cite{parismitasarma2023} can be accessed through hyperlink \\
	(\url{https://data.mendeley.com/datasets/dtvbwrhznz/4}).
	
	\item Self-curated dataset by the authors is available upon request.
\end{itemize}

\bibliographystyle{splncs04}
\bibliography{sn-bibliography}
\end{document}